\documentclass[journal,twoside,web]{ieeecolor}
\usepackage{generic}
\usepackage{cite}
\usepackage{amsmath,amssymb,amsfonts}
\usepackage{algorithmic}
\usepackage{graphicx}
\usepackage{algorithm,algorithmic}
\usepackage[dvipsnames]{xcolor}
\usepackage[hidelinks]{hyperref}
\usepackage{textcomp}
\def\BibTeX{{\rm B\kern-.05em{\sc i\kern-.025em b}\kern-.08em
    T\kern-.1667em\lower.7ex\hbox{E}\kern-.125emX}}
\begin{document}
\title{Tabular LLMs for Interpretable Few-Shot Alzheimer's Disease Prediction with Multimodal Biomedical Data}
\author{Sophie Kearney$^*$ and Shu Yang$^*$, Zixuan Wen, Weimin Lyu, Bojian Hou, Duy Duong-Tran, Tianlong Chen, Jason H. Moore, Marylyn D. Ritchie, Chao Chen, Li Shen$^\dag$, for the Alzheimer’s Disease Neuroimaging Initiative$^{**}$
\thanks{$^*$ These authors contributed equally to this research.}
\thanks{**Data used in the preparation of this article were obtained from the Alzheimer’s  Disease Neuroimaging Initiative (ADNI) database (adni.loni.usc.edu). As such, the investigators within the ADNI  contributed to the design and implementation of  ADNI  and/or provided data but did not participate in the analysis or writing of this report.  A  complete  listing  of  ADNI  investigators  can  be found at: \url{http://adni.loni.usc.edu/wp-content/uploads/how_to_apply/ADNI_Acknowledgement_List.pdf}}
\thanks{S. Kearney, S. Yang, Z. Wen, B. Hou, D. Duong-Tran, and L. Shen are with the University of Pennsylvania, Philadelphia, PA, USA
(e-mail: sophie.kearney@pennmedicine.upenn.edu; yangsh@pennmedicine.upenn.edu; zxwen@sas.upenn.edu; houbo@pennmedicine.upenn.edu; duyanh.duong-tran@pennmedicine.upenn.edu; li.shen@pennmedicine.upenn.edu).}
\thanks{C. Chen and W. Lyu are with Stony Brook University, Stony Brook, NY, USA (e-mail: chao.chen.1@stonybrook.edu; welyu@cs.stonybrook.edu)}
\thanks{T. Chen is with the University of North Carolina at Chapel Hill, Chapel Hill, NC, USA
(e-mail: tianlong@cs.unc.edu).}
\thanks{M. Ritchie is with the Medical University of South Carolina, Charleston, SC, USA (email: ritchiem@musc.edu)}
\thanks{J. H. Moore is with Cedars-Sinai Medical Center, West Hollywood, CA, USA
(e-mail: jason.moore@csmc.edu).}
\thanks{$^\dag$L. Shen is the corresponding author.}}

\maketitle

\begin{abstract}
Accurate diagnosis of Alzheimer's disease (AD) requires handling tabular biomarker data, yet such data are often small and incomplete, where deep learning models frequently fail to outperform classical methods. 
Pretrained large language models (LLMs) offer few-shot generalization, structured reasoning, and interpretable outputs, providing a powerful paradigm shift for clinical prediction.
We propose \textbf{TAP-GPT}, \textit{\underline{T}abular \underline{A}lzheimer's \underline{P}rediction \underline{GPT}}, a domain-adapted tabular LLM framework built on TableGPT2 and fine-tuned for few-shot AD classification using tabular prompts rather than plain texts.
We evaluate TAP-GPT across four ADNI-derived datasets, including QT-PAD biomarkers and region-level structural MRI, amyloid PET, and tau PET for binary AD classification.
Across multimodal and unimodal settings, TAP-GPT improves upon its backbone models and outperforms traditional machine learning baselines in the few-shot setting while remaining competitive with state-of-the-art general-purpose LLMs. 
We show that feature selection mitigates degradation in high-dimensional inputs and that TAP-GPT maintains stable performance under simulated and real-world missingness without imputation.
Additionally, TAP-GPT produces structured, modality-aware reasoning aligned with established AD biology and shows greater stability under self-reflection, supporting its use in iterative multi-agent systems. 
To our knowledge, this is the first systematic application of a tabular-specialized LLM to multimodal biomarker-based AD prediction, demonstrating that such pretrained models can effectively address structured clinical prediction tasks and laying the foundation for tabular LLM-driven multi-agent clinical decision-support systems.
The source code is publicly available on GitHub: \url{https://github.com/sophie-kearney/TAP-GPT}.

\end{abstract}


\section{Introduction}
\label{sec:Introduction}

The transition toward precision health in the management of complex human diseases such as Alzheimer's disease (AD) calls for the sophisticated investigation of heterogeneous, multimodal biomedical data.
As a progressive and irreversible neurodegenerative disorder, AD represents the most common form of dementia globally and the sixth leading cause of mortality in the United States~\cite{PRINCE201363, 2022e105}.
Currently, curative treatments for AD remain limited, posing a critical challenge for both clinical neuroscience research and public health, especially driven by the accelerated aging of the global population. 
Consequently, accurate and early detection becomes essential for proactive intervention.
Previous studies have identified a set of AD biological markers and verified their ability to measure AD pathology. 
The importance of these biomarkers was subsequently reflected in revised diagnostic criteria. 
In 2018, the National Institute on Aging (NIA) and the Alzheimer’s Association Workgroup’s Research Framework proposed to use the AT[N] biomarker classification system \cite{JACK2018535, jack2016t}. 
By defining AD in terms of a biological, rather than a clinical, construct, this system divides the major AD biomarkers into three categories based on the nature of their pathology: 1) ‘A’: the $\beta$-amyloid biomarkers (as measured by amyloid positron emission tomography (PET) imaging or cerebrospinal fluid (CSF) A$\beta_{42}$); 2) `T': the pathological tau biomarkers (CSF phosphorylated Tau or Tau PET imaging); and 3) `N': the neurodegeneration or neuronal injury biomarkers (atrophy on structural magnetic resonance imaging (MRI) scans, Fluorodeoxyglucose (FDG) PET scans, or CSF total Tau) \cite{jack2016t}. 
The multifaceted AT[N] biomarkers have been widely utilized in AD research and investigated in conjunction with other types of markers like cognitive measures for disease progression \cite{MOSCONI2007129}. 

From a data‑science perspective, these AT[N] measurements are often represented as tables: 
each subject corresponds to a row, each measurement (e.g., CSF tau biomarker or brain MRI-derived regional feature) corresponds to a column.
In the real-world setting, row/sample sizes of  biomedical tables are frequently modest relative to the column/feature dimensionality, and missingness is common during the data collection.
Such tabular structure creates distinctive modeling challenges. 
Unlike language or images, tables lack a natural ordering or locality (e.g., columns are permutation invariant~\cite{hollmann2022tabpfn, dash-etal-2022-permutation}), and predictive signals may reside in cross-feature interactions that are difficult to capture with small size cohort. 
As a result, conventional machine learning methods, such as gradient‑boosted and tree‑ensemble approaches, have remained strong baselines for many small datasets.
In particular, for AD outcome prediction, Random Forest (RF) and XGBoost have shown robust performance by modeling complex, non-linear relations among features from tabular data~\cite{bao2024combined, yi2023xgboost, guo2025machine}; even logistic regression (LR) remains a clinical standard due to its simplicity and clarity~\cite{li2025prediction}.
In contrast, classic deep neural networks can be sensitive to limited sample sizes, feature scaling, and missing values~\cite{grinsztajn2022tree,mcelfresh2023neural}, which are commonly seen in AD tabular setting~\cite{ShenThompson2020, Baoj2024, lin2021deep}.
Despite the predictive performance, these conventional machine learning and deep learning methods provide limited insight into which blood marker or brain region drives a given prediction, and they do not leverage prior knowledge in the biomedical domain beyond the training data.

Recently, transformer-based large foundation models offer an alternative paradigm and open up new opportunities for handling AD tabular data: rather than training a predictive model from scratch, one can adapt a pretrained model that has already learned broad prior knowledge over data distributions and reasoning patterns~\cite{fang2024large, badaro2023transformers}.
This paradigm includes both: \textit{(i)} tabular foundation models that learn in-context prediction behaviors for tables~~\cite{hollmann2022tabpfn, hollmann2025accurate, zhangmitra, matabdpt, balazadehcausalpfn}, and \textit{(ii)} large language model-based approaches that convert tables into textual prompts through serialization or latent embeddings through table encoders~\cite{zhang2023generative, wangchain, han2024large, li2024cancergpt, hegselmann2023tabllm, lee2025knowledge}.
Such methods can conduct few-shot prediction for very small datasets and incorporate broad prior knowledge, potentially improving performance on biomedical tables like AD AT[N] markers.
Tabular foundation models (TFMs) tailored to small-to-medium datasets have emerged~\cite{zhangmitra, matabdpt, balazadehcausalpfn}. 
As an example, TabPFN~\cite{hollmann2022tabpfn, hollmann2025accurate} represents one of the most significant works in this class. 
It performs in-context learning (ICL) to predict labels from a small training table in a single forward pass without gradient updates. 
Although ICL was first observed in LLMs~\cite{brown2020language}, later studies showed transformers can leverage ICL to approximate classical learning algorithms such as logistic regression and neural networks~\cite{zhou2023algorithms, muller2021transformers}. 
Pre-trained via prior-fitting on 100 million synthetic datasets spanning diverse causal structures, TabPFN achieved state-of-the-art accuracy on small, noisy tables with outliers and missing values, requiring only ~50\% of the data to match conventional models~\cite{hollmann2022tabpfn, hollmann2025accurate}. 
Notable extensions have explored adapting TabPFN to larger or more complex tasks through finetuning and retrieval-augmentation~\cite{thomas2024retrieval, wu2025efficient}.
However, TFMs are primarily predictive engines and do not natively provide natural-language reasoning that is valuable for clinical interpretability, error analysis, and downstream agentic workflows.

In parallel, pretrained on large text corpora, LLMs exhibit emergent abilities for tabular tasks, such as tabular data generation, table understanding, prediction, reasoning and even feature selection/engineering, \textit{etc}. \cite{zhang2023generative, wangchain, han2024large, li2024cancergpt, hegselmann2023tabllm, lee2025knowledge}. 
With prompting and finetuning, they can be adapted to tabular tasks, addressing limitations of naive transformer architectures for handling tabular formats. 
In particular, as the leading open-source efforts of the field, TableGPT unifies tables, language, and functional commands for table understanding and manipulation with natural language instructions \cite{zha2023tablegpt}.
It utilizes dedicated table encoders and compresses tabular information into compact latent representations to mitigate token-length constraints and capture schema-level context.
TableGPT2~\cite{su2024tablegpt2} scales this idea via large-scale pretraining on 593K tables and 2.36M table-question pairs, with a schema-aware table encoder and a 7B–72B parameter QWen2.5 LLM decoder \cite{hui2024qwen2}.
It achieved $\sim$35–49\% improvements over base LLMs on diverse table-related business intelligence tasks.
The most recent variant of the family, TableGPT-R1, released in December 2025~\cite{yang2025tablegpt}, goes beyond standard supervised fine-tuning by using a systematic reinforcement learning approach and a powerful QWen3~\cite{yang2025qwen3technicalreport} backbone to strengthen multi-step reasoning and advanced table understanding tasks~
\cite{yang2025tablegpt}.
Despite these strong gains from tabular LLMs, studies show that they still lag behind humans on complex, real-world tables that require deep domain knowledge~\cite{su2024tablegpt2, wu2025tablebench}.

Building on these advances, we introduce a domain-adapted tabular LLM framework for AT[N]-based AD diagnosis, termed \textit{TAP-GPT: Tabular Alzheimer's Prediction GPT}, which novelly adapts the multimodal TableGPT2 model for few-shot classification of AD vs cognitively normal (CN) individuals from AT[N] feature tables. 
We formulate the task as tabular few-shot in-context learning paradigm, where the model is prompted with a small table containing several example subjects with neuroimaging, \textit{etc.}, features and diagnosis labels, alongside a target subject whose label to be predicted.
To adapt TableGPT2 for AD tabular prediction, we employ diverse strategies to construct dedicated training tables from clinical data, and we perform parameter-efficient finetuning~\cite{dettmers2023qlora} on these tables. 
To our knowledge, this represents the first systematic application of tabular LLM to AD biomarker study, uniquely combining tabular-LLM's table understanding and reasoning capabilities with specific medical knowledge regarding AD and AT[N] markers.
We evaluated TAP-GPT against a set of general-purpose LLMs and TFM, as well classic machine learning methods, effectively demonstrating the potential of TAP-GPT for structured biomedical prediction and paving the way for future multi-agent diagnostic systems on multimodal AD data.
A preliminary version of TAP-GPT has been reported in the ACM-BCB conference in 2025~\cite{10.1145/3765612.3767229}. 
We applied TAP-GPT on the QT-PAD biomarker dataset derived from ADNI (cohort 1, GO, 2), with 15 clinical markers and 4 covariates as features. 
In this study, we present a substantial expansion of the conference proceeding by comprehensively exploring and enhancing TAP-GPT, and we make the following key contributions here:

\begin{itemize}
    \item \textbf{Generalization:} Validated model generalization by adapting TAP-GPT to three additional neuroimaging datasets: amyloid PET, tau PET, and structural MRI, with regional measures stored in tabular format. Extended beyond the single QT-PAD dataset for few-shot AD prediction. 
    \item \textbf{Enhanced Interpretability:} Enhanced interpretability analyses with multimodal reasoning and self-reflection prompting, laying the foundation for future multi-agent AD diagnosis workflows.
    \item \textbf{Feature Selection:} Developed  systematic feature selection for high-dimensional tables via classic LASSO and LLM-guided ranking approaches, under context‑length constraints. 
    \item \textbf{Robustness to Missingness:} Conducted robustness analysis under both simulated and naturally occurring missing data derived from ADNI.
\end{itemize}

\section{Methods}
\label{sec:Methods}

We start this section by introducing our few-shot and in-context learning notions \textbf{in the context of tabular prediction tasks} following previous works~\cite{zha2023tablegpt, su2024tablegpt2, hollmann2022tabpfn, hollmann2025accurate, thomas2024retrieval, wu2025efficient}, as they are slightly different from the normal non-tabular scenario with LLMs.

Formally, let $\mathcal{M}$ denote the set of modalities considered in this study, including QT-PAD clinical biomarkers, structural MRI, amyloid PET, and tau PET. 
For each modality $m \in \mathcal{M}$, we define a modality-specific dataset 
$D^{(m)} = \{(x^{i,(m)}, y^i)\}_{i=1}^{N_m}$ consisting of $N_m$ labeled subject samples.
Each sample $i$ in modality $m$ is represented by a feature vector 
$x^{i,(m)} = (x^{i,(m)}_1, \dots, x^{i,(m)}_{d_m}) \in \mathbb{R}^{d_m}$ 
containing $d_m$ structured biomarkers or regional imaging measurements, and a corresponding clinical diagnosis label 
$y^i \in \{0, 1\}$, where $0$ denotes cognitively normal (CN) and $1$ denotes Alzheimer's disease (AD), consistent across QT-PAD and imaging datasets. 
For each modality $m$, the table columns (features) are described by a set of natural language feature names 
$F^{(m)} = \{f^{(m)}_1, \dots, f^{(m)}_{d_m}\}$, which are incorporated directly into the prompt construction. 

\begin{figure}[h]
  \centering
  \includegraphics[width=\linewidth]{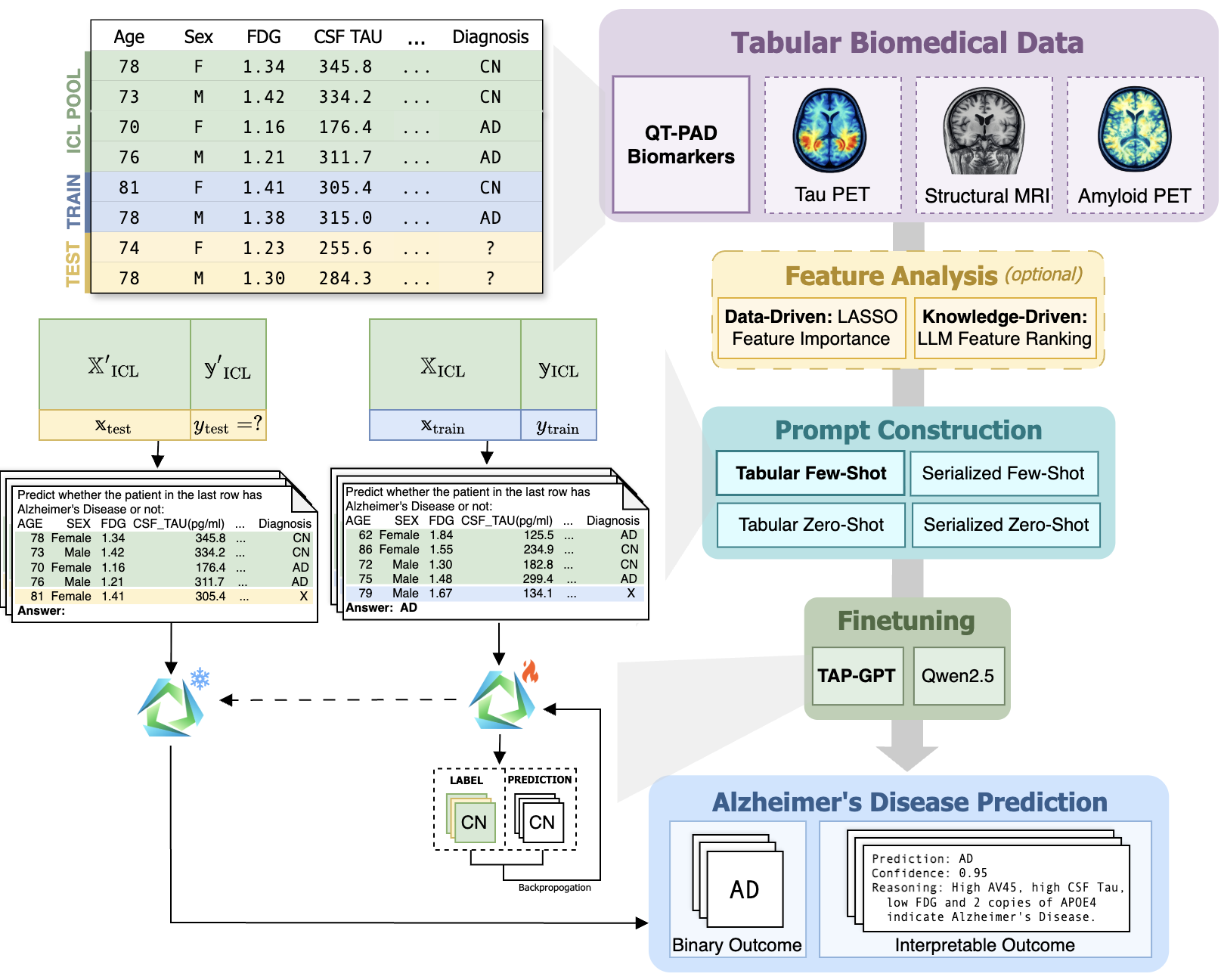}
  \caption{Overview of the TAP-GPT framework. For each dataset, we split the subjects into training, testing, and pools for in-context examples, with an optional feature selection step. We construct tables for finetuning and evaluating TAP-GPT for the task of Alzheimer's disease prediction. }
  \label{fig:overview}
\end{figure}

The goal of the tabular prediction task for modality $m$ is to learn a model that, given a new subject's feature vector 
$x^{(m)}_{\text{test}} \in \mathbb{R}^{d_m}$ 
and (optionally) contextual examples in the form of a structured table, can accurately predict the corresponding diagnosis 
$y_{\text{test}} \in \{0, 1\}$. 
While this work focuses on binary classification, the formulation naturally generalizes to tabular regression or multiclass classification tasks. An overview of this framework is provided in Figure~\ref{fig:overview}.

In the few-shot setting, prediction for a target patient $x_t^{(m)}$ is conditioned on a set of $k$ labeled in-context examples drawn from a split-specific, disjoint in-context learning (ICL) pool. For each data split $s \in \{\text{train}, \text{val}, \text{test}\}$, we define a corresponding ICL pool $\mathcal{I}_s^{(m)}$ that is used exclusively to construct prompts for samples in that split. Formally, for modality $m \in \mathcal{M}$, the context set for a target sample in split $s$ is defined as
$\mathcal{C}_t^{(m)} = \{(x_{t_j}^{(m)}, y_{t_j})\}_{j=1}^{k},$
where each $(x_{t_j}^{(m)}, y_{t_j})$ is sampled from the same split’s ICL pool $\mathcal{I}_s^{(m)}$ and does not overlap with the target sample. The ICL pools are used solely for prompt construction and are not directly used as supervision outside their corresponding split.

A structured prompt is constructed by combining the in-context examples with the unlabeled target patient:
$P_t^{(m)} = \Phi\big(\mathcal{C}_t^{(m)}, x_t^{(m)}\big),$
where $\Phi(\cdot)$ denotes a deterministic formatting function that renders the examples and target patient into either a tabular representation (preserving explicit column structure) or a serialized natural language representation.

TAP-GPT is an autoregressive language model parameterized by $\theta$ that defines a conditional distribution over output tokens given the prompt: $p_\theta(y_t \mid P_t^{(m)}).$ 
The model is fine-tuned to maximize the likelihood of the correct diagnosis label under this conditional distribution. During inference, we restrict the output space to the binary label set $\{0,1\}$ using constrained decoding. The predicted diagnosis is therefore obtained as
$\hat{y}_t = \arg\max_{y \in \{0,1\}} 
p_\theta(y \mid P_t^{(m)})$.


To predict $\hat{y}_t$, some prior approaches rely on simple serialization strategies that convert structured tabular data into plain-text sequences before prompting pretrained LLMs~\cite{hegselmann2023tabllm, lee2025knowledge}. In such methods, a serialization function  $\mathtt{Serialize}(x^{(m)}, y, F^{(m)})$ 
uses the feature names $F^{(m)}$ to transform each row into a natural language description (e.g., ``The$f_1$ biomarker value is $x_1$.''), then a generic LLM directly processes the resulting text to generate a prediction: $\hat{y}_t = \mathtt{LLM}(\mathtt{Serialize}(\mathcal{C}_t^{(m)}, x_t^{(m)}))$.
However, this approach loses explicit column alignment and structured relationships present in the original table, and can become token-inefficient as the number of features or in-context examples increases~\cite{hegselmann2023tabllm}.

In contrast, our proposed framework, TAP-GPT, makes use of TableGPT2's encoder--decoder architecture with a semantic table encoder, which maintains structural awareness, and an LLM decoder.
Let $g_{\theta_0}$ denote the pretrained TableGPT2 model, originally trained on large-scale generic tabular data for table understanding tasks.

To adapt this model to supervised Alzheimer’s disease prediction, we perform task-specific fine-tuning using QLoRA. Let $\Delta\theta$ denote low-rank parameter updates applied to the decoder weights. The adapted parameters are therefore $\theta = \theta_0 + \Delta\theta.$ In our framework, the pretrained semantic table encoder is kept fixed, preserving its learned structural representations of tabular data. Low-rank adapters are injected into the decoder, allowing the model to specialize to the binary diagnostic objective while updating only a small subset of parameters. The resulting modality-specific predictor $f_\theta^{(m)}$ is trained to approximate the conditional distribution $p(y \mid x^{(m)}, \mathcal{C}^{(m)})$,
by maximizing the likelihood of the correct diagnosis label given the structured prompt, without updating its weights during inference~\cite{muller2021transformers}. In the following subsections, we provide more details for the data, framework and experimental setup.

\subsection{Datasets}

Data used in the preparation of this article is obtained from the Alzheimer's Disease Neuroimaging Initiative (ADNI) database (\url{http://adni.loni.usc.edu}) \cite{weiner2013alzheimer,weiner2017recent}. The ADNI was launched in 2003 as a public-private partnership led by Principal Investigator Michael W. Weiner, MD. The primary goal of ADNI has been to test whether serial MRI, PET, other biological markers, and clinical and neuropsychological assessment can be combined to measure the progression of mild cognitive impairment (MCI) and early AD. 
All participants provided written informed consent, and study protocols were approved by each participating site’s Institutional Review Board (IRB). Up-to-date information about the ADNI is available at \url{www.adni-info.org}.


\subsubsection{Quantitative templates for the progression of AD Biomarker Data}

The first dataset we focus on is a subset of the ADNI 1/Go/2 cohorts, i.e.,  the Quantitative templates for the progression of AD (QT-PAD) project dataset \footnote{\url{https://www.pi4cs.org/qt-pad-challenge}}, which was initially developed as an AD Modeling Challenge from ADNI and has been widely used for diverse AD-related research. The QT-PAD dataset contains multimodal biomarkers indicating progression of AD across the AT[N] framework and beyond. We excluded the five congnitive score metrics because these measurements are heavily used by clinicians for diagnosis, and are therefore highly collinear to diagnosis. The resulting dataframe contained 15 clinical markers for each subject: PET measures (FDG PET and Amyloid PET), cerebrospinal fluid measures (CSF ABETA, CSF TAU, CSF PTau), structural MRI-derived measures from FreeSurfer (Whole brain, hippocampus, entorhinal cortex, ventricles, mid-temporal lobe, fusiform gyrus), as well as one genetic marker (APOE4 status) and three demographic covariates including age, gender and years of education. We removed any samples with missing values and extracted the baseline measurement of each patient which resulted in 237 CN and 96 AD patients, 333 in total.

\subsubsection{Regional Summary Imaging Data}

The second dataset we focus on is a subset of participants from the ADNI-1, ADNI-GO, and ADNI-2 cohorts with available amyloid PET, tau PET, and structural MRI regional summary data. Amyloid-$\beta$ and Tau PET images were processed by the UC Berkeley PET Analysis Group using standardized preprocessing pipelines. The extracted regional PET standardized uptake value ratios (SUVRs) using FreeSurfer-defined cortical and subcortical regions of interest (ROI). Structural MRI volumetric measures were also extracted, yielding region-level volumetric features aligned with PET data. 

As features, we used region-level summaries derived from FreeSurfer, including 68 cortical regions (34 per hemisphere) and four subcortical regions (thalamus, hippocampus, amygdala, and caudate), resulting in 74 regional summaries across all three modalities. In addition, we included three demographic covariates (age, gender and years of education) and one genetic marker, copies of APOE4. 

\begin{figure}[h]
  \centering
  \includegraphics[width=0.75\linewidth]{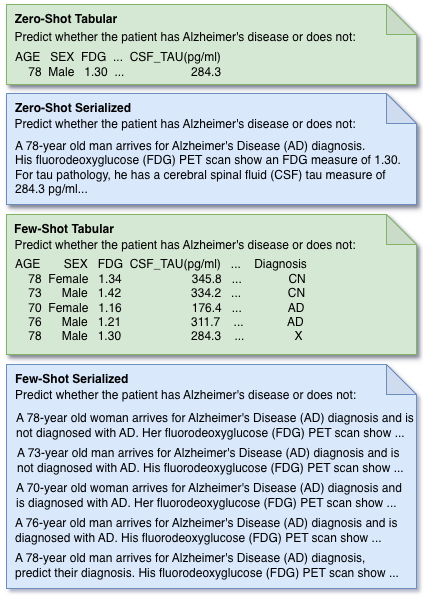}
  \caption{The four prompt formats used in our experiments, with tabular prompts shown in green and serialized prompts in blue. In each case, the model is asked to predict Alzheimer’s disease status for the same held-out patient. All values shown are synthetic with abbreviated prompts. }
  \label{fig:prompt_formats}
\end{figure}

For each modality, we selected the most recent available visit to maximize the number of AD cases and improve class balance, restricting the cohort to participants with either CN or AD diagnoses. In the amyloid PET modality, this resulted in 1,031 patients (691 CN, 340 AD) and the tau PET modality, this resulted in 610 patients (484 CN, 126 AD). Structural MRI data were available for participants in both PET cohorts, so to maximize sample size, we extracted MRI features from the amyloid PET cohort, resulting in the same set of 1,031 participants.

\subsubsection{Creating Tabular Prompts}

To support in-context learning, we designed a splitting strategy that strictly separates model training, validation, and testing data from samples used for prompt construction. For each dataset (QT-PAD, amyloid PET, tau PET, and structural MRI), we allocated approximately 40\% of samples to training, 10\% to validation, and 20\% to testing. The remaining data were divided into three non-overlapping ICL pools corresponding to the train, validation, and test splits (approximately 10\% each). ICL pools are used exclusively to populate prompts and are never included in model optimization or evaluation.

We designed four prompt formats for the binary classification task to evaluate performance under different contextual and representational settings (Figure~\ref{fig:prompt_formats}). In each case, we construct a patient-level prompt and the model is asked to predict whether the patient is cognitively normal or has AD. The formats vary along two dimensions: contextual exposure (zero-shot versus few-shot) and input representation (tabular versus serialized natural language). In all settings, we append the patient information to a standardized natural language instruction requesting a binary classification for the target sample.

\begin{itemize}
    \item \textbf{Zero-Shot Tabular} A single unlabeled row in a tabular format (one target sample).
    \item \textbf{Zero-Shot Serialized} A single unlabeled natural language description of one patient's information.
    \item \textbf{Few-Shot Tabular} For each unlabeled target patient, \textit{k} labeled examples are randomly sampled from the corresponding ICL pool. These \textit{k} rows are presented above the target patient in the table, with the target appearing in the final row and their diagnosis label masked.
    \item \textbf{Few-Shot Serialized} Mirroring the few-shot tabular setting, but each patient’s features are converted into natural language. The \textit{k} in-context examples include diagnosis labels, while the target sample is presented without a label.
\end{itemize}

For serialized prompts, we convert each patient's structured features into a standardized natural language description while preserving consistent formatting across datasets. Pronouns are adjusted according to recorded sex, and measurement units are retained when applicable. For imaging modalities, we present patient-level covariates first, followed by regional imaging features formatted as structured key–value pairs (ROI=value) to preserve clarity across numerous measurements. In contrast, QT-PAD biomarkers, which span multiple biological domains, are assembled into a clinically interpretable narrative summary.
We display tabular prompts with the \textit{head()} function from the pandas library. Column order is consistent between all data splits.
We embedded all formats of presenting patient data in a natural language prompt instructing the model to provide a diagnosis with instructions tailored to the prompt format.

For interpretability experiments, we modified only the instruction and expected output format while keeping patient data and in-context examples identical to the corresponding non-interpretable prompts. We appended \textit{``Let’s think step by step.’’} and instructed the model to return a JSON object containing (i) the predicted label (\texttt{0}=CN, \texttt{1}=AD), (ii) a free-text reasoning explanation, and (iii) a scalar confidence score. The supervised objective remained binary classification, enabling interpretable outputs without altering the training loss or architecture.

\subsection{TAP-GPT Framework}

We present TAP-GPT, a domain-adapted tabular large language model build on TableGPT2-7B \footnote{\url{https://huggingface.co/tablegpt/TableGPT2-7B}}, that integrates a static tabular encoder with an LLM decoder backbone (Qwen2.5-7B). While TableGPT2 was originally optimized for business-oriented tabular understanding tasks, we repurpose its decoder for structured clinical prediction of Alzheimer’s disease. Specifically, we finetune the decoder using labeled prompts that contain small tables of biomedical features, enabling the model to map structured patient data to diagnostic outcomes (Figure~\ref{fig:overview}).

For each dataset and random seed, we generate tabular inputs for every patient in the training, testing, and validation test splits using their corresponding ICL pools. For each patient, we convert this structured table into four prompt formats (zero-shot tabular, zero-shot serialized, few-shot tabular, and few-shot serialized) and finetune a separate TAP-GPT model for each format to isolate the impact of structure and context. 

We train TAP-GPT using supervised learning to predict a binary diagnosis label (AD=1, CN=0). We place the diagnosis token in a fixed position within each prompt to ensure consistent parsing and evaluation across seeds and formats. We repeat all experiments across ten random seeds (36, 73, 105, 254, 314, 492, 564, 688, 777, and 825).

To adapt the large decoder, we apply parameter-efficient finetuning using QLoRA implemented with the HuggingFace transformers, peft, and bitsandbytes libraries. QLoRA updates low-rank adaptation matrices while keeping the base model weights quantized in 4-bit precision, which reduces memory requirements without full weight retraining. Due to architectural constraints of the released TableGPT2 model, we keep the tabular encoder fixed and update only the decoder parameters during adaptation. This design allows us to preserve general-purpose tabular representations while injecting domain-specific diagnostic knowledge into the LLM.

\begin{figure*}[t]
  \centering
  \includegraphics[width=0.7\textwidth]{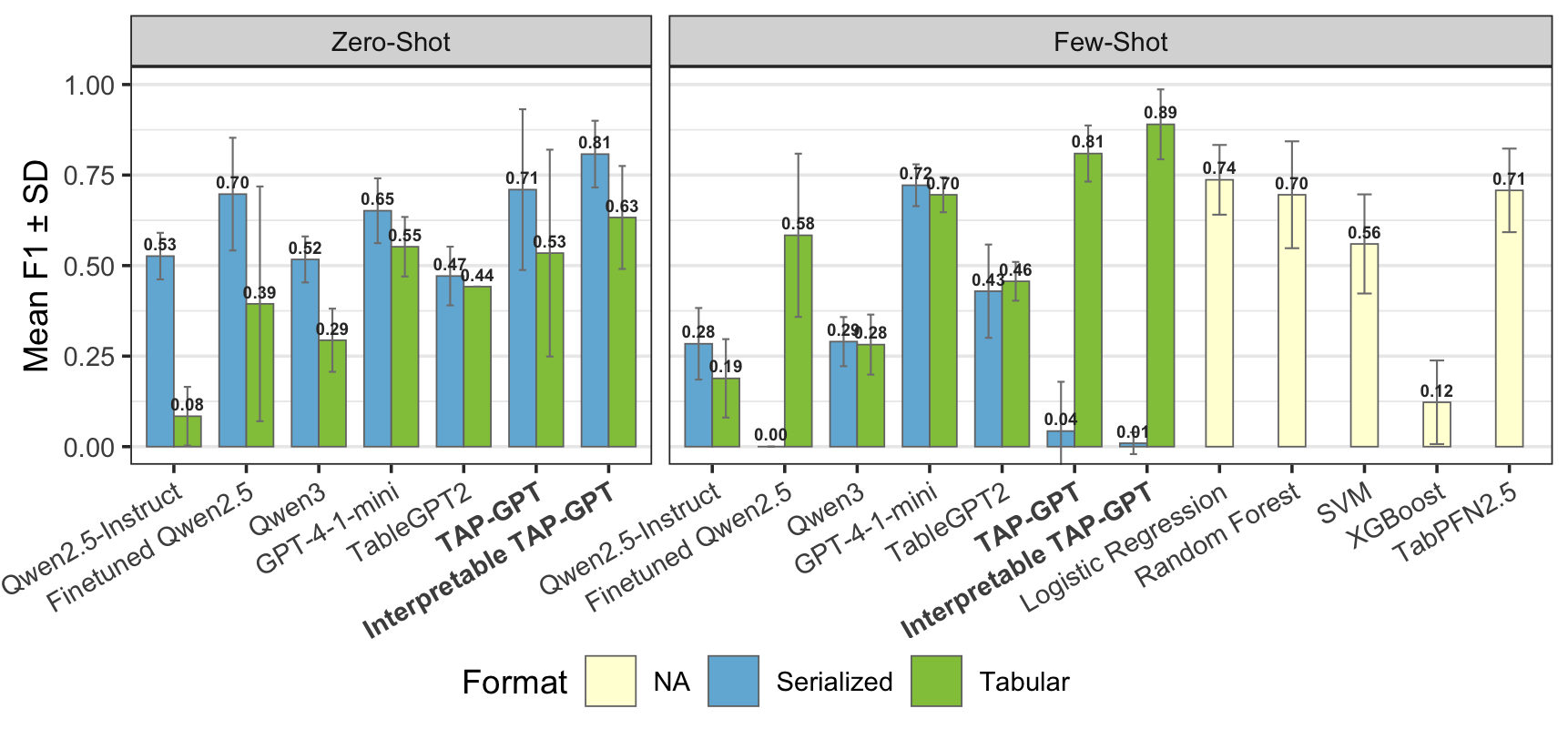}
  \caption{QT-PAD mean F1 across models in zero-shot and few-shot ($k=8$) contexts. LLMs use tabular (green) and serialized (blue) prompts with error bars for standard deviation; TabPFN and traditional ML (yellow) operate directly on structured data.}
  \label{fig:qtpad_results}
\end{figure*}

During inference, we constrain the model’s output space to \{0,1\} using a custom logits processor to enforce valid binary predictions. For interpretability experiments, we modify the prompts to additionally request a structured reasoning explanation and a scalar confidence score, while maintaining the same supervised objective of binary classification. This prompt-level modification enables interpretable outputs without altering the underlying training loss or architecture.


\subsection{Experimental Setup}

We designed a set of experiments to evaluate TAP-GPT across four multimodal biomedical datasets. For LLM-based models, we evaluated four prompt formats, as shown in Figure \ref{fig:prompt_formats}. Only the model type and prompt configuration varied across experiments. Performance was primarily assessed using F1 score, with balanced accuracy, precision, and recall also reported.

For the QT-PAD tabular dataset, we performed a \textit{k}-ablation study on the validation set to determine the optimal number of in-context examples, where \textit{k} denotes the number of in-context examples. We finetuned TAP-GPT models separately for each 
$k \in \{2,4,6,8,10,12,16,20\}$ across three seeds under the tabular few-shot prompt format. We used validation F1 to select the \textit{k} that achieved the most stable and consistently high performance across seeds and used this value for all subsequent few-shot tabular experiments.

Given the substantially larger feature spaces in the imaging-derived datasets (MRI, amyloid PET, tau PET), we incorporated feature selection prior to prompt construction to mitigate performance degradation due to prompt length. Using the training set only, we applied LASSO to select the top-$p$ features per modality and seed. We conducted a joint $p \times k$ ablation for $p \in \{8,16,32\}$ and $k \in \{4,8,12\}$ to characterize trade-offs between feature dimensionality and in-context supervision. Optimal configurations were selected based on validation F1 and applied to final test evaluation.

To benchmark TAP-GPT, we compared performance across all prompt formats and datasets with a diverse set of baselines. We included unfinetuned TableGPT2 to assess the impact of task-specific finetuning and incorporated TableGPT-R1 for imaging analyses. To isolate the contribution of tabular specialization, we evaluated Qwen2.5-7B-Instruct (the backbone of TableGPT2) and Qwen3-8B as general-purpose LLMs. We also included GPT-4.1-mini as a strong general LLM benchmark for zero- and few-shot inference. To examine the effect of finetuning without tabular pretraining, we finetuned Qwen2.5-7B-Instruct under identical training conditions. We included TabPFN as a non-LLM tabular foundation model, tested only on few-shot tasks. Finally, we benchmarked four traditional machine learning models on the few-shot tasks: Logistic Regression, Random Forest, XGBoost, and Support Vector Machine—on the few-shot tabular setting to provide classical baselines.

For every dataset, we performed hyperparameter tuning using Optuna to maximize validation F1. For finetuned TableGPT2 and Qwen2.5 models, we tuned LoRA rank, dropout rate, learning rate, batch size, weight decay, maximum training steps, and learning rate scheduler. 
For imaging experiments with feature selection, we treated the number of features\textit{p} as a tunable parameter within the\textit{p} $\times$ \textit{k}ablation grid.
We tuned traditional machine learning models over their respective standard hyperparameters (e.g., regularization strength, tree depth, kernel type, and subsampling strategies). For each model and modality, we selected the configuration achieving consistently strong validation F1 across seeds for final test evaluation.

We conducted all experiments on a high-performance computing cluster using NVIDIA A100 GPUs (80 GB HBM memory) with 160 GB system RAM per node. We allocated each training job a single GPU, 4 CPU cores, and 1 node under a SLURM scheduler. Due to the large number of experimental configurations (modalities, seeds, prompt formats, and \textit{p} $\times$ \textit{k}combinations), we executed several hundred GPU jobs. We stored model checkpoints and adapters via Hugging Face Hub, with approximately 0.25 TB of model storage accumulated across 335 experiment variants. This large-scale compute infrastructure enabled systematic evaluation across modalities and ablation settings while maintaining strict separation of training, validation, and test data.
More details of this study, such as LLMs' hyperparameters, various prompts, together with the source code, are all publicly available in our GitHub repository: \url{https://github.com/sophie-kearney/TAP-GPT}.

\section{Results}
\label{sec:Results}


\subsection{Overall Model Performance}

\begin{figure*}[t]
  \centering
  \includegraphics[width=0.9\textwidth]{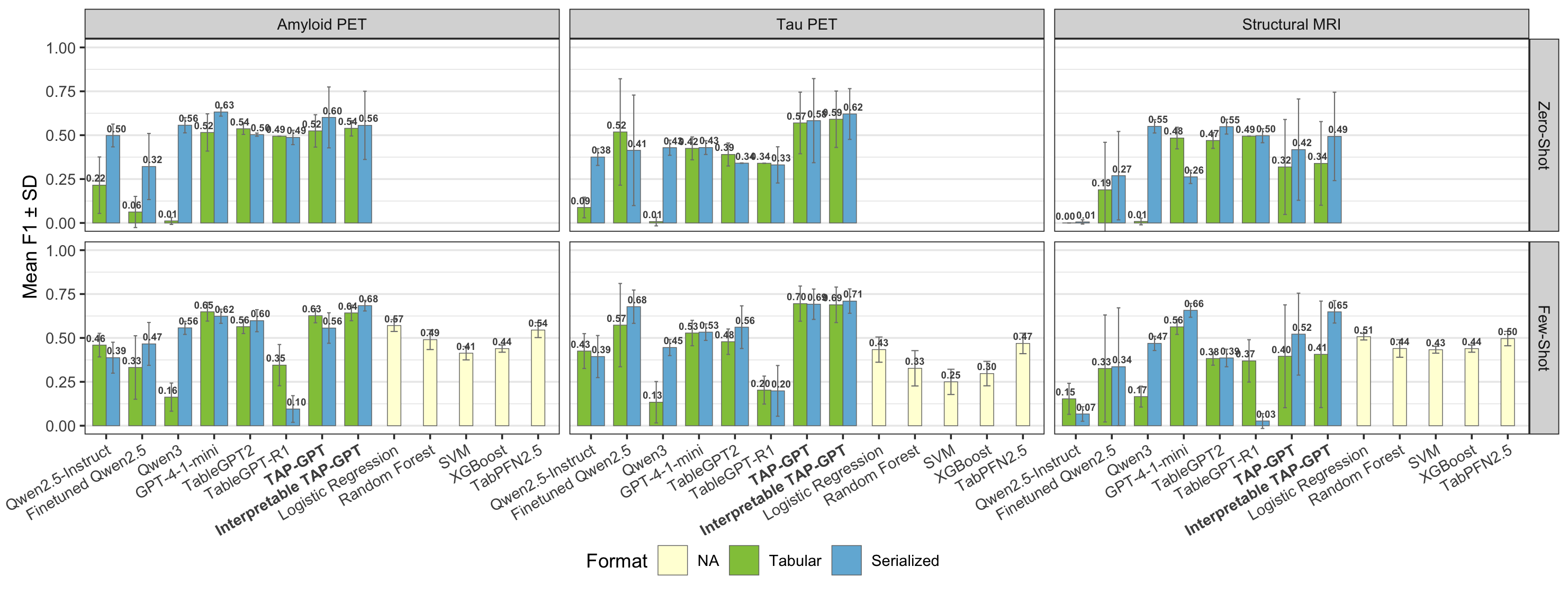}
  \caption{Imaging ROI mean F1 across models in the zero- and few-shot ($k=4$) contexts. Tabular and Serialized prompts are given to LLMs, with traditional ML and TabPFN models directly using the data. Model performance is evaluated across three imaging modalities, Amyloid PET, Tau PET, and Structural MRI.}
  \label{fig:imaging_results}
\end{figure*}

We evaluated TAP-GPT performance on the test set across four prompt formats (zero-shot tabular, zero-shot serialized, few-shot tabular, and few-shot serialized). To contextualize performance, we evaluated TAP-GPT against four model categories: (1) classical machine learning baselines (logistic regression, random forest, SVM, XGBoost) (2) TabPFN to represent state-of-the-art tabular foundation model approaches; (3) vanilla LLMs without task-specific fine-tuning, including Qwen2.5-Instruct, Qwen3-Instruct, and GPT-4.1 mini as a strong external LLM from a different model family; and (4) tabular-specialized derivative TableGPT2. Because TAP-GPT is fine-tuned from TableGPT (itself adapted from Qwen2.5-Instruct), this hierarchy allows us to assess the incremental contribution of tabular specialization and disease-specific fine-tuning. Interpretable TAP-GPT is the modified prompting for interpretable output from the same TAP-GPT model.

\subsubsection{Model Performance on Biomarker Data}
  

In the zero-shot setting (Figure \ref{fig:qtpad_results}, left), serialized prompts consistently outperformed tabular prompts. Because the tabular zero-shot format contains only a single patient row without in-context examples, the serialized representation provides a more natural linguistic structure for encoding clinical information. 
TAP-GPT achieved strong results in both settings, with stronger results under serialization. Notably, introducing prompting for interpretable outputs with CoT reasoning produced a substantial improvement over the standard TAP-GPT configuration in both tabular and serialized formats. The interpretable TAP-GPT variant achieved the highest overall zero-shot performance.

TAP-GPT and Interpretable TAP-GPT have the strongest performance in the few-shot tabular context (Figure \ref{fig:qtpad_results}, right), likely due to pairing the tabular understanding of TableGPT with task-specific knowledge acquired from finetuning. Incorporating CoT interpretable responses again boosted TAP-GPT's performance in this task to mean F1=0.89. As expected, TabPFN2.5 performs strongly in this context despite the very low-shot setting. Notably, TAP-GPT outperforms all traditional machine learning baselines, TabPFN, and vanilla LLMs in the few-shot tabular task. GPT-4.1-mini performs well across both formats in this context, which suggests that updating the backbone of TAP-GPT with more capable base models could further improve prediction performance. In contrast, performance in the serialized few-shot setting is substantially weaker across most LLMs, with several models collapsing toward majority-class predictions. This degradation may reflect the increased prompt length and reduced structural clarity of serialized inputs when combined with multiple in-context examples.

\subsubsection{Model Performance on Imaging Data}




We next evaluated TAP-GPT on the unimodal imaging-derived datasets amyloid PET, tau PET, and structural MRI volume under a 4-shot setting with 16 selected features. We tested the same models on these datasets as with the QT-PAD data, with the addition of  TableGPT-R1, the reasoning version of TableGPT, built on Qwen3 backbone.


The classic machine learning models can only be tested in the few-shot setting and are not subject to prompt formatting because they operate directly with the data. In this very low-shot setting of four training samples and one test sample, these methods naturally struggle to perform competitively to the LLMs, which benefit from semantic knowledge and pre-training. However, TAP-GPT consistently outperforms all traditional methods and TabPFN.

In the few-shot tabular setting, across each of the datasets, we observe a consistent performance progression from the base backbone to the domain-adapted model. For example, in the tau modality, Qwen2.5 started with a mean F1 of 0.43, then TableGPT2 had a mean F1 of 0.48, and then TAP-GPT showed an F1 of 0.68. A similar stepwise trend is reflected in the amyloid modality, and although not as strong, in the MRI modality.

Unlike the QT-PAD dataset, the serialized prompts (blue in Figure \ref{fig:imaging_results}) perform similarly to the tabular prompts (green). In this setting, we implemented serialization using a structured key-value format appended to the minimal patient description, rather than fully narrative text. Because key–value serialization preserves much of the original structural organization and does not substantially inflate prompt length, it remains closer to a semi-structured representation. As a result, performance between tabular and serialized formats remain similar in this unimodal imaging setting, in contrast to the larger format-driven differences observed in QT-PAD.

In contrast to the QT-PAD results, the strongest comparator in the imaging-derived setting is GPT-4.1 mini, a more advanced general-purpose LLM. In the tabular few-shot setting, TAP-GPT performs similarly or better across modalities. For amyloid PET, Interpretable TAP-GPT achieves an F1 of 0.64, comparable to GPT-4.1 mini (F1 = 0.65). In the tau modality, TAP-GPT reaches an F1 of 0.68 compared to 0.53 for GPT-4.1 mini. In the MRI modality, GPT-4.1 mini achieves the highest tabular few-shot performance (F1 = 0.66). While TAP-GPT’s tabular performance was lower in this modality, the serialized Interpretable TAP-GPT achieves competitive performance (F1 = 0.65), closely matching GPT-4.1 mini.

For general-purpose LLMs, the shift from zero-shot to few-shot prompting did not produce the larger gains observed in QT-PAD. However, few-shot prompting consistently improved TAP-GPT performance relative to zero-shot. A similar trend is observed for GPT-4.1 mini. These results indicate that, even in unimodal settings where strong prior knowledge may already exist in pretrained models, access to labeled in-context examples provides additional task-specific signal.

\subsection{Ablations}

\begin{figure}[h]
  \centering
  \includegraphics[width=\linewidth]{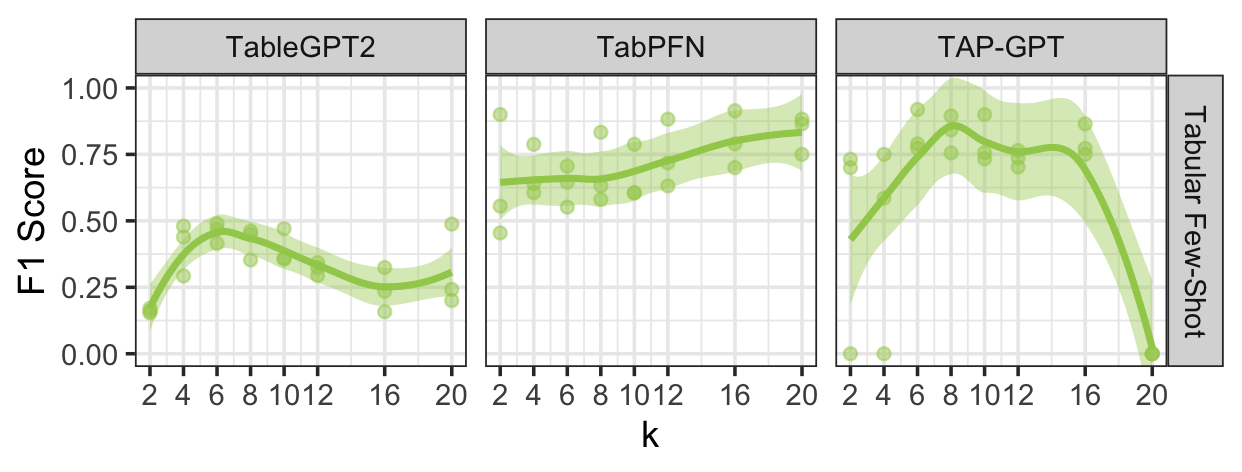}
  \caption{Number of ICL examples (\textit{k}) ablation analysis on QT-PAD data across TableGPT2, TabPFN, and TAP-GPT. TabPFN performance steadily improved with larger k, TableGPT2 improved up to $k=6$ and declined thereafter, and TAP-GPT peaked at $k=8$.}
  \label{fig:k_ablation}
\end{figure}

\subsubsection{\textit{k}-Ablation Analysis for Biomarker Data}

To choose an optimal number of ICL examples ($k$) in the few-shot setting, we tested a set of different $k$s $\{2,4,6,8,10,12,16,20\}$ on TAP-GPT, TableGPT2, and TabPFN (Figure \ref{fig:k_ablation}). Testing a $k$ value on TAP-GPT requires re-finetuning TableGPT for that format of prompt across all seeds, so due to GPU memory and sample size restrictions, we restricted the upper limit of $k$ to 20 and tested across three seeds. TableGPT performance seems to peak around $k=6$ to $k=8$ but remains relatively steady beyond this point. As expected because it is built for much larger tables than this few-shot setting, TabPFN shows increasing performance with a higher $k$. TAP-GPT has the highest mean F1 at $k=8$ (0.831, SD = 0.0702), with $k=6$ and $k=10$ achieving similar performance. We selected $k=8$ for all few-shot experiments with QT-PAD data.

\begin{figure}[h]
  \centering
  \includegraphics[width=\linewidth]{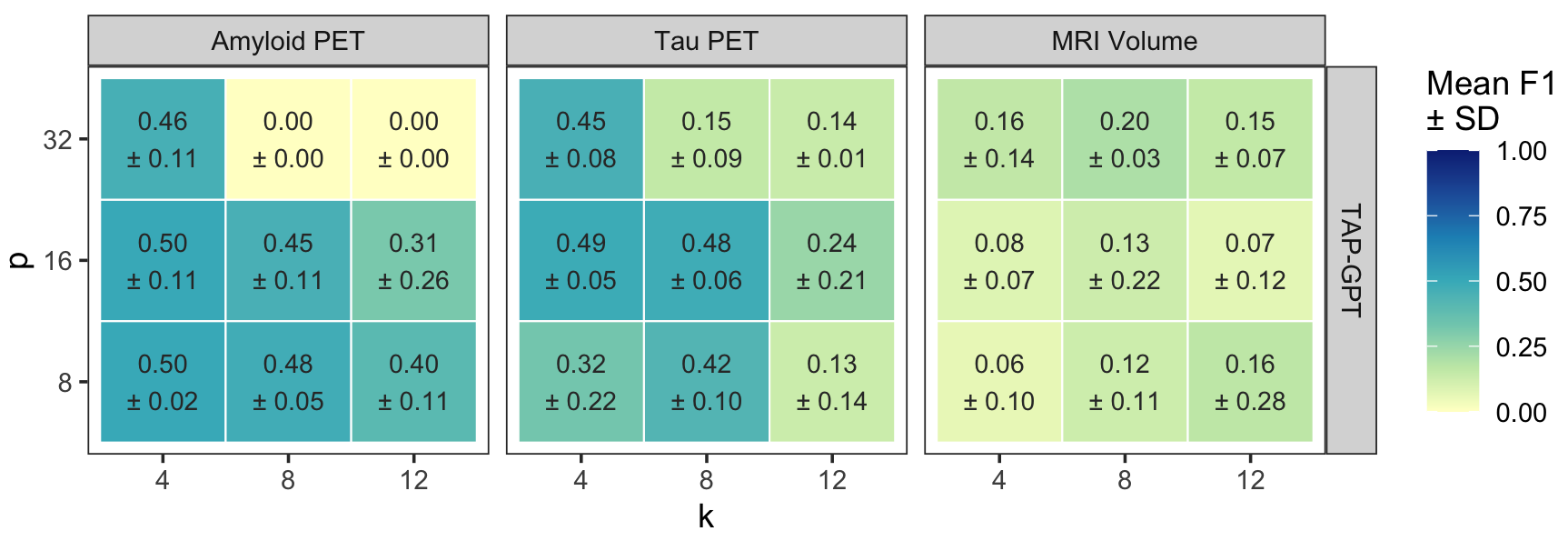}
  \caption{Number of ICL examples (\textit{k}) and number of features (\textit{p}) ablation analysis on imaging data for TAP-GPT. Mean F1 score (± standard deviation across three random seeds) is shown for each (\textit{k}, \textit{p}) combination.
  }
  \label{fig:pxk}
\end{figure}

\subsubsection{\textit{k}- and \textit{p}- Ablation Analysis for Imaging Data}

To determine the appropriate values for the number of ICL examples (\textit{k}) and selected features (\textit{p}) for few-shot prompting, we conducted an ablation study for the tabular few-shot prompt on the TAP-GPT model for each modality. The correct choice of (\textit{k}, \textit{p}) balances prompt length and necessary informational content to optimize performance. For each modality, \textit{p} denotes the number of imaging-derived regional features selected from the full set of 68 cortical and 4 subcortical FreeSurfer regions, while patient-level covariates (APOE4 status, age, sex, and years of education) were always included and not counted toward \textit{p}. We finetuned TableGPT2 on tabular few-shot prompts with each combination of $p \in (8,16, 32)$ and $k \in (4, 8, 12)$ across three arbitrarily chosen seeds (36, 73, 314) for each modality dataset, as seen in Figure \ref{fig:pxk}.

Performance across amyloid PET and tau PET consistently peaked at lower values of $k$ and $p$, further highlighting the limitation of tabular LLMS with large tables as text input. $k=4$ shows the most stability across values of $p$ for amyloid and tau modalities and pairs best with moderate feature counts of $p=16$. In contrast, the MRI volumetric modality showed weaker signal overall with some inconsistent and variable gains at larger $p$ or $k$ values. To maintain consistency across modalities, we set $p=16$ and $k=4$ for all imaging modality prompts. In addition, $p = 16$ provides a feature dimensionality similar to the QT-PAD biomarker dataset, enabling more consistent comparison of TAP-GPT performance across all datasets evaluated in this study.

\begin{figure}[h]
  \centering
  \includegraphics[width=\linewidth]{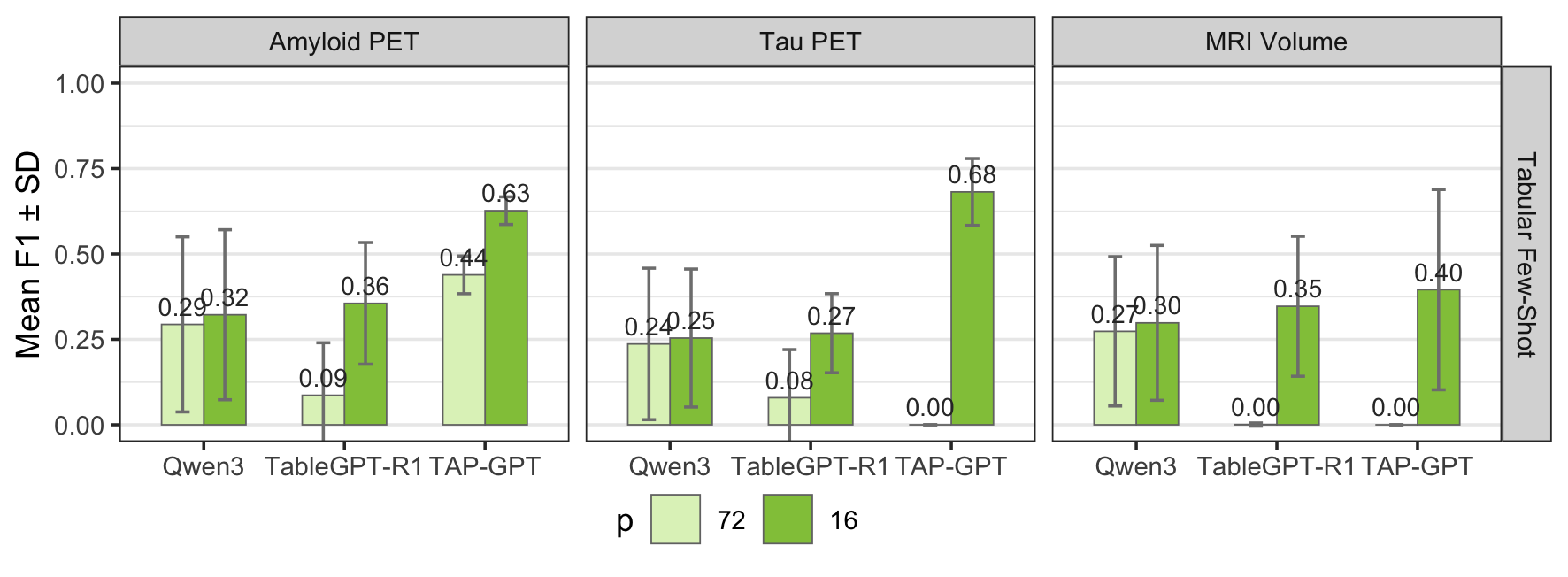}
  \caption{TAP-GPT Performance using all 72 ROIs compared to selecting the top 16 features for each modality.}
  \label{fig:no_feature_selection}
\end{figure}

\subsubsection{Impact of Feature-Dimensionality on Tabular LLMs}

Figure \ref{fig:no_feature_selection} examines the impact of feature dimensionality on few-shot tabular performance across amyloid PET, tau PET, and MRI, comparing TAP-GPT to two related baselines, Qwen3-8B and TableGPT-R1. For each modality, we evaluate performance using either all 72 available regional features ($p = 72$; light green) or a reduced subset of the top 16 features selected via LASSO ($p = 16$; dark green). This experiment isolates the effect of feature selection in a long-context setting and assesses whether newer foundation models better tolerate high-dimensional tabular inputs.

Across all three modalities, TAP-GPT shows consistent improvement in performance when feature selection is applied. In contrast, using all 72 features leads to performance degradation, particularly for tau PET and MRI, supporting the use of feature selection for high-dimensional tables for tabular LLMs. While Qwen3 largely maintains comparable performance between $p = 72$ and $p = 16$, its overall performance remains lower than TAP-GPT. TableGPT-R1 exhibits sensitivity to feature dimensionality, with near-zero performance in the $p = 72$ setting across modalities, indicating difficulty handling long tabular contexts without feature reduction.

These results suggest improvements in base model architecture may not be sufficient to address long-context challenges with high-dimensional clinical tables. Feature selection enables effective performance for tabular LLMs.

\begin{figure}[h]
  \centering
  \includegraphics[width=\linewidth]{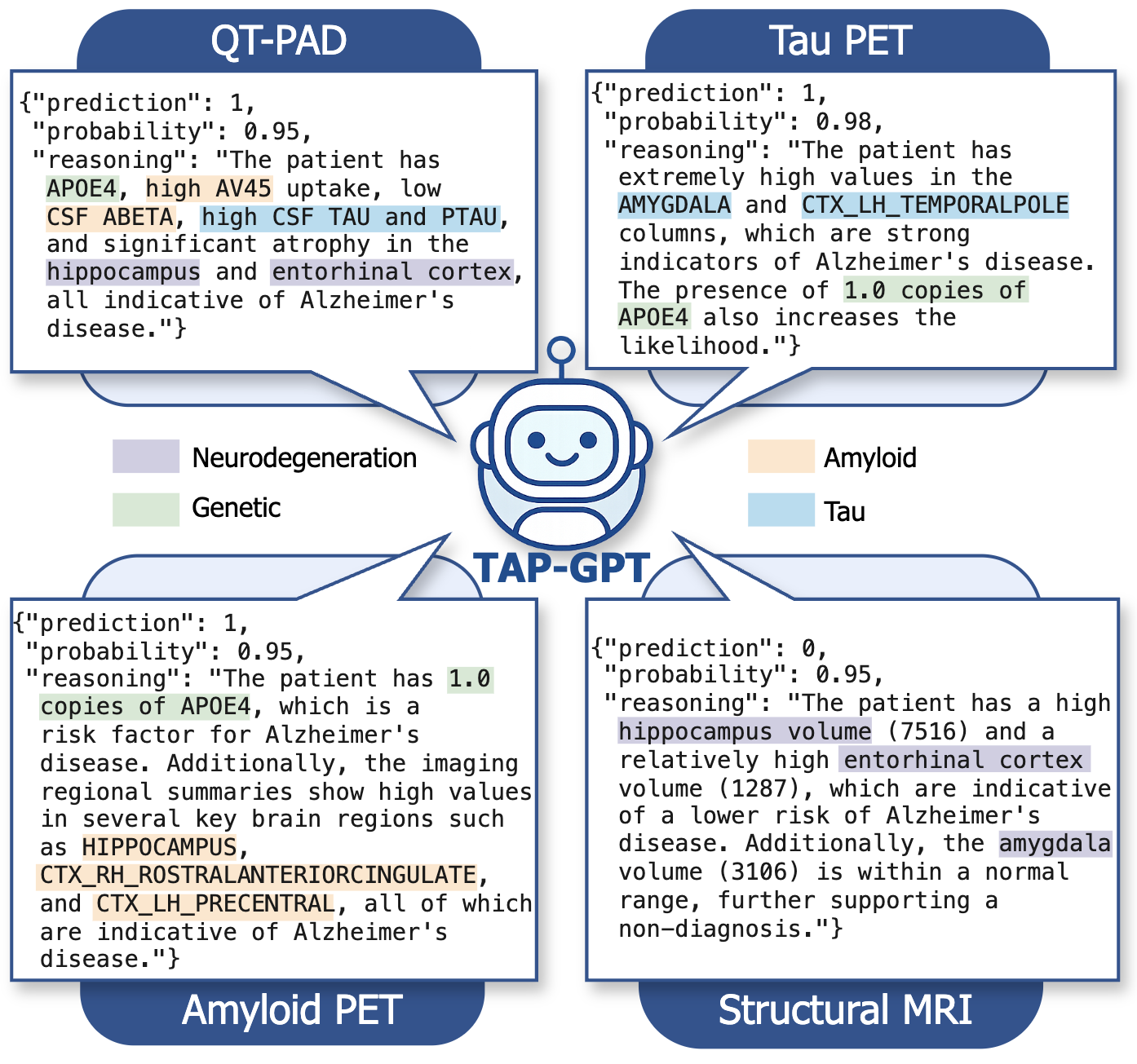}
  \caption{Example interpretable outputs from TAP-GPT across all datasets. For each modality, the model correctly predicts the patient's diagnosis and returns a structured JSON containing the prediction, confidence score, and reasoning. 
  }
  \label{fig:interpretability}
\end{figure}

\subsection{Interpretability}

\subsubsection{Interpretable Prompting Reveals Multimodal Reasoning}

In addition to improving predictive performance, interpretable prompting provides direct insight into how TAP-GPT synthesizes multimodal clinical features (Figure \ref{fig:interpretability}). Across all four datasets, TAP-GPT produces structured JSON outputs containing the binary prediction, a probability estimate, and a stepwise reasoning explanation. By qualitatively inspecting the outputs, we noticed modality-specific feature prioritization. 

In the QT-PAD dataset, TAP-GPT often references features spanning multiple biological domains when explaining its predictions, including genetic risk factors, patterns in amyloid and tau biomarkers, and measures of neurodegeneration. This cross-domain reasoning suggests that the model integrates multimodal inputs to form a more comprehensive patient profile.


For the imaging modalities, even when restricted to the selected subset of 16 features, TAP-GPT emphasizes regions consistent with established Alzheimer’s disease biology. In the structural MRI example shown in Figure \ref{fig:interpretability}, the model correctly interprets higher volumetric values as preserved structure rather than increased atrophy and cites normal hippocampal and amygdala volumes to support the classification of cognitively normal. In the tau PET and amyloid PET examples, the model references APOE4 status alongside modality-specific regional uptake patterns, combining genetic risk with imaging evidence in its reasoning. These examples illustrate how finetuning enables TAP-GPT to leverage both domain-specific training signals and the underlying semantic knowledge encoded in the base LLM to generate clinically interpretable explanations.

The interpretability framework also provides flexibility in output structure. Because interpretability is induced via prompt design rather than architectural modification, we can request alternative structured outputs (e.g., predicted AD risk probability, ranked important features, or feature-level justifications) without retraining the model. This adaptability enables downstream integration into clinical decision-support pipelines where explanation format may vary by use case.

At the same time, our qualitative analysis reveals reasoning inconsistencies. In some cases, the model misinterprets feature directionality (e.g. treating elevated FDG values as indicative of AD when reduced metabolism is typically associated with disease) or incorrectly references values from other rows as belonging to the target patient. These errors underscore the need to validate generated explanations and caution against interpreting reasoning strings as causal evidence. Exposing the model’s intermediate reasoning provides feedback that can guide targeted refinement of prompting strategy or additional finetuning. The ability to surface and inspect model reasoning provides additional diagnostic signal unavailable in conventional tabular classifiers.

\subsubsection{LLM-Derived Feature Ranking Across Imaging Modalities}
\begin{figure}[h]
  \centering
  \includegraphics[width=0.9\linewidth]{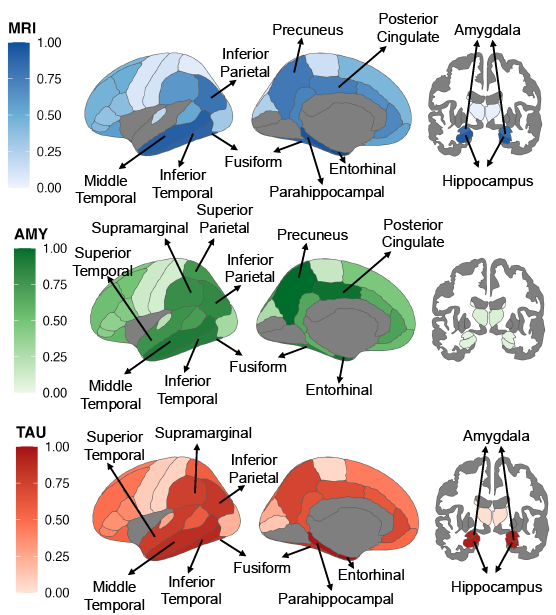}
  \caption{Regional feature importance as ranked by GPT-4.1-mini across modalities. Cortical and subcortical regions are colored according to their normalized importance scores (0–1), derived from the LLM-generated ranking of input features. Labeled regions reflect the top 10 ROIs chosen for each modality.
  }
  \label{fig:gpt_ranking_map}
\end{figure}

We prompted GPT-4.1-mini to rank each of the ROIs found in the imaging datasets according to their importance in diagnosing AD for each modality and mapped these normalized rankings onto anatomical space in Figure \ref{fig:gpt_ranking_map}. We used GPT-4.1-mini because TAP-GPT is finetuned on a subset of features, and GPT-4.1-mini performs strongly on this task.

In the MRI modality, GPT-4.1-mini prioritized several medial temporal and subcortical structures, including the hippocampus and amygdala, alongside posterior cortical areas such as the precuneus and posterior cingulate. Many of these regions are established in neurodegeneration and were also selected by LASSO, demonstrating overlap between LLM-derived and data-driven feature selection.
In the amyloid modality, the model emphasized posterior cortical regions but placed less weight on frontal areas than both prior literature and the LASSO-based selection, suggesting a different pattern of importance across cortical lobes. 
For tau PET, GPT-4.1-mini concentrated importance in medial temporal and inferior temporal regions, including the entorhinal cortex and parahippocampal areas, aligning with the known spatial progression of tau pathology in AD. 

Despite these anatomically plausible patterns, GPT-4.1-mini consistently selected a partially distinct subset of regions compared to LASSO across modalities. These differences indicate that the LLM may rely on broader clinical and biological knowledge rather than purely on statistical sparsity in the training data, supporting the use of LLM-derived rankings as an alternative or complementary feature selection strategy. GPT-4.1-mini could act as a helper agent to create an optimized set of features prior to prompt creation, operating in a similar knowledge space as tabular LLM predictor agents.

\subsubsection{Self-Reflection}

\begin{figure}[h]
  \centering
  \includegraphics[width=\linewidth]{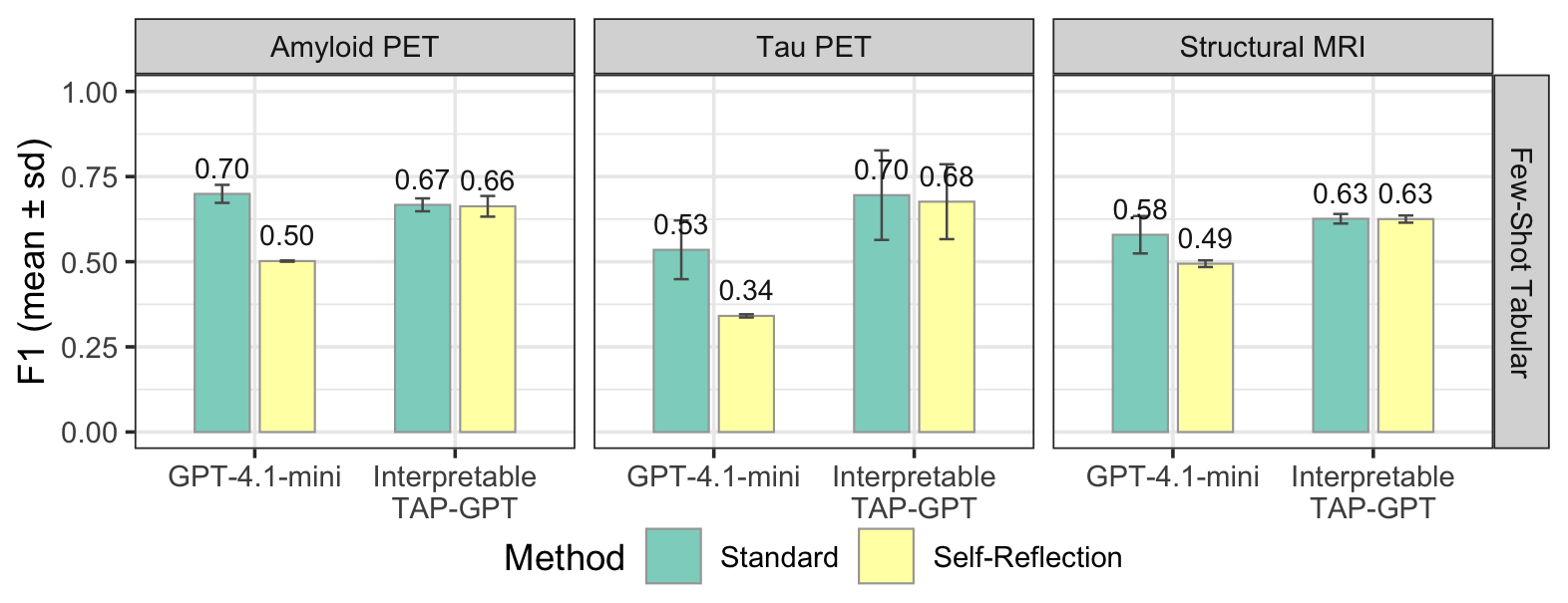}
  \caption{Effect of self-reflection on GPT-4.1-mini and TAP-GPT across modalities. We compare standard prompting to a self-reflection setting in which each model reviews and potentially revises its initial prediction.
  }
  \label{fig:self_reflection}
\end{figure}

To evaluate stability under iterative reasoning, we introduced a self-reflection step where both TAP-GPT and GPT-4.1-mini were prompted to review and optionally revise their own predictions across three arbitrarily selected seeds (688, 825, and 492). Figure \ref{fig:self_reflection} shows that GPT-4.1-mini shows considerable performance loss under self-reflection across modalities. In contrast, TAP-GPT maintains similar performance between standard and self-reflection conditions across all three imaging modalities. Although both models achieve comparable performance in the standard few-shot tabular setting, TAP-GPT demonstrates significantly greater prediction stability when subjected to iterative prompting.

Consistency under self-reflection is important for multi-agent system (MAS) deployment, where models engage in multiple rounds of critique, discussion, and voting. A candidate model for MAS must tolerate iterative reasoning without destabilizing its predictions. Although this represents a simple single round of self-reflection, TAP-GPT's stability here indicates reduced susceptibility to drift during debate.

\subsection{Missingness}

We evaluated the robustness of TAP-GPT to missing data in two contexts on QT-PAD biomarker dataset: (i) controlled simulated missingness applied uniformly across the input table, and (ii) real-world missingness presented in held-out clinical samples. 

\begin{figure}[h]
  \centering
  \includegraphics[width=\linewidth]{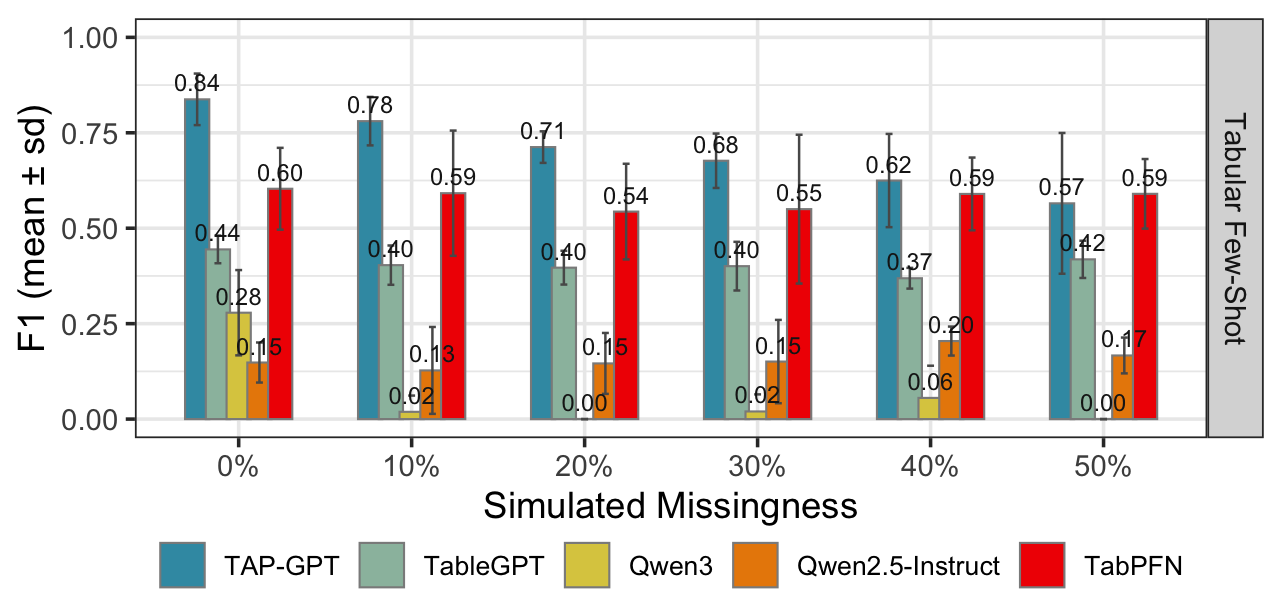}
  \caption{{Performance of multiple models on simulated missingness in the QT-PAD dataset at inference. TAP-GPT retains high performance until high missingness at 50\%.}
  }
  \label{fig:masked_missingness}
\end{figure}

\subsubsection{Performance Under Simulated Missingness}
In the simulated missingness setting, we randomly masked 10\%, 20\%, 30\%, 40\%, and 50\% of feature values across the entire input table provided to each model under a missing-completely-at-random assumption. As shown in Figure \ref{fig:masked_missingness}, TAP-GPT  maintained a clear performance advantage over baseline tabular LLMs and classical models as missingness increased. Performance declined gradually with increasing masking, but TAP-GPT remained competitive through 40\% missingness. At the most extreme setting (50\% missingness), TabPFN achieved performance comparable to TAP-GPT, while other baselines degraded substantially. These results suggest that TAP-GPT is robust to moderate levels of uniformly distributed missing data and that tool-calling TabPFN may be appropriate in extreme cases of missingness.

\begin{figure}[h]
  \centering
  \includegraphics[width=\linewidth]{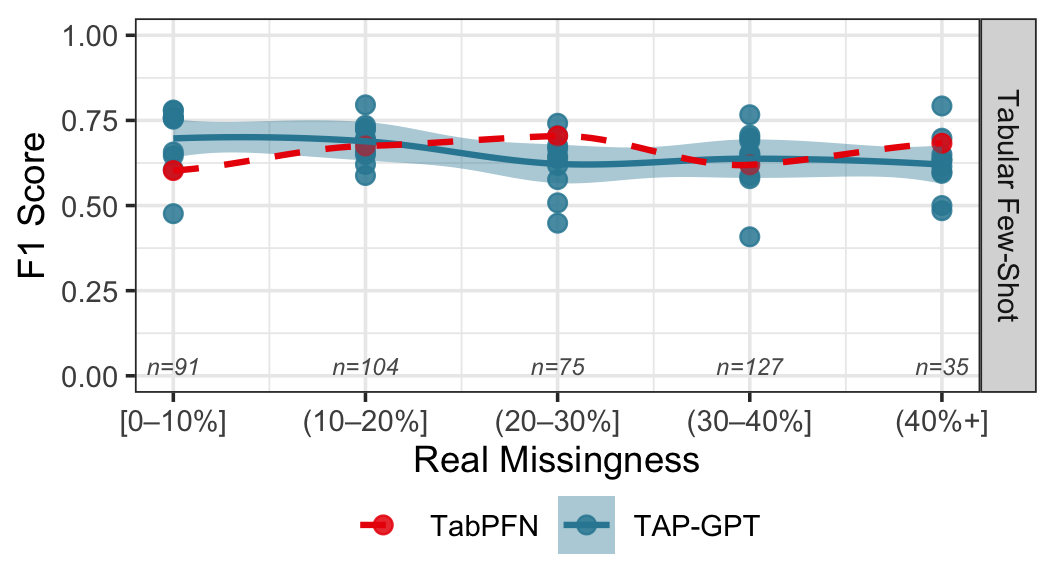}
  \caption{Performance across real-world target missingness levels. est samples were grouped by the proportion of missing features in the target patient only, while the in-context learning (ICL) pool retained its natural missingness distribution across all patients with missing data (mean missingness = 0.247).
  }
  \label{fig:real_missingness}
\end{figure}

\subsubsection{Performance Under Naturally Occurring Missingness}
To assess robustness under realistic clinical conditions, we next evaluated TAP-GPT on naturally occurring missingness. Patients from QT-PAD with missing values were excluded from both training and prior evaluation of TAP-GPT and therefore provide an independent cohort for this analysis. From this subset, 541 patients with naturally occurring missing data were identified, and 20\% were reserved exclusively to serve as an ICL pool, leaving 433 target patients. For each target patient in the evaluation set, tabular prompts were constructed by randomly sampling $k=8$ examples from this pool, which has an average missingness of 0.247. Missingness is defined for one patient as $\frac{\text{number of missing features}}{\text{number of total features}}$. The ICL examples reflect the natural, heterogeneous missingness present in QT-PAD and were not stratified by missingness level. All 10 TAP-GPT models, each trained on a different random seed from the fully observed cohort, were evaluated on the same set of target patients to characterize robustness across independently trained models.

Figure \ref{fig:real_missingness} stratifies results by target-level missingness, isolating the effect of missing data in the patient being evaluated while preserving a deployment-like prompting scenario. TAP-GPT exhibits stable performance across missingness bins, with limited degradation and relatively low variability across seeds. TabPFN maintains similar performance to TAP-GPT across increasing missingness.

\section{Discussion}
\label{sec:Discussion}

In this work, we present TAP‑GPT, a domain‑adapted tabular‑LLM framework for few‑shot AD prediction from multimodal biomedical tables, and a comprehensive analysis of its ability. 
In biomedical settings, where datasets are low-sample, partially observed, and multimodal, conventional deep learning methods often fail to outperform tree-based models. Our work introduces a new paradigm for structured clinical prediction by adapting tabular-specialized large language models to multimodal Alzheimer's disease biomarker and imaging-derived data. To our knowledge, this is the first systematic application of tabular LLMs to multimodal biomedical biomarkers and imaging-derived modalities, showing that structured LLM architectures can bridge prior knowledge, in-context learning, and domain adaptation to address long-standing challenges in clinical tabular modeling.

Across experiments spanning one multimodal biomarker dataset (QT-PAD), and three unimodal imaging-derived datasets, TAP-GPT shows strong and consistent performance in low-shot settings. On the QT-PAD dataset, TAP-GPT achieves particularly strong results in the few-shot tabular setting, outperforming vanilla LLMs, traditional machine learning models, and TabPFN.
On the imaging datasets, TAP-GPT shows competitive performance across prompt settings, although in some configurations, strong general-purpose LLMs achieve similar results. GPT-4.1-mini and Qwen3 are both stronger models than the backbone of TAP-GPT, Qwen2.5, but we see that in most cases TAP-GPT keeps up with or surpasses these models. It is possible that moving towards a multimodal analysis of the imaging data would further improve TAP-GPT's performance on these datasets, similar to what we observed in the multimodal dataset QT-PAD, which benefits from multiple biological domains. 

Our results also highlight the importance of the underlying LLM backbone. Even without task-specific finetuning, GPT-4.1-mini achieved strong performance on the QT-PAD data, and even outperformed TAP-GPT in some imaging experiments. This underscores the growing tabular reasoning capacity of high-capacity general-purpose LLMs. 
Within our framework, however, we observe consistent gains from Qwen2.5 to TableGPT2 through tabular specialization, and from TableGPT2 to TAP-GPT with task-specific fine-tuning, indicating that each stage of specialization contributes meaningfully to performance.
Although TableGPT-R1, built on Qwen3, did not perform strongly in our few-shot imaging experiments (and we similarly did not observe strong performance from Qwen3 in this setting) these findings reinforce that backbone selection remains critical. Notably, we did not enable extended “thinking” or reasoning modes for these models, as our objective was constrained to binary classification.
At the same time, the fixed architecture of TableGPT2 limits our ability to replace its decoder with newer, more capable LLMs. Future work could explore reconstructing a similar tabular-LLM architecture using more modern backbones for clinical prediction tasks.

An important finding is that TAP-GPT remains competitive with TabPFN in the low-shot settings common in biomedical research. TabPFN is specifically designed for few-shot tabular prediction and sets a strong baseline for small structured datasets. In the QT-PAD dataset with 4–10 examples (Figure \ref{fig:k_ablation}), TAP-GPT performs comparably while also providing interpretable outputs. In the 4-shot imaging experiments, TAP-GPT often outperforms TabPFN, depending on the prompt format. Unlike TabPFN, which relies solely on tabular structure, TAP-GPT combines a tabular encoder with a language model decoder, allowing it to use both structural patterns and prior semantic knowledge. This may offer an advantage in clinical settings, where labeled data are scarce and feature names carry meaningful domain information.

Another practical advantage of LLM-based prediction models is their flexibility in handling missing data. Unlike traditional machine learning models or tabular foundation models that typically require explicit imputation, TAP-GPT can represent missingness directly within the prompt without artificially filling values. In our experiments, TAP-GPT maintained stable performance under both simulated and real-world missingness, remaining competitive up to moderate levels of missing data and showing limited degradation across target-level missingness bins. These results suggest that LLM-based models can tolerate incomplete clinical tables without heavy preprocessing, which is especially important in real-world biomedical settings where data are often sparse or irregular.

Beyond performance, several properties of TAP-GPT position it as a candidate agent for clinical multi-agent systems. First, its robustness to both real-world and simulated missingness in the multimodal setting suggests it can operate under the incomplete and heterogeneous data conditions typical to clinical workflows, where different agents may receive partial views of a patient’s record. Second, the model produces interpretable outputs that explicitly reference multimodal evidence, demonstrating that it performs cross-modal reasoning rather than relying on isolated signals. Importantly, these explanations can be generated in flexible natural language or constrained structured formats, making them suitable for both clinician-facing interpretation and machine-readable communication between agents. Third, while GPT-4.1-mini achieved comparable or stronger performance in some settings, TAP-GPT exhibited greater stability under self-reflection, indicating more consistent reasoning across iterative evaluations. This stability is particularly relevant for multi-agent chains of communication, where models repeatedly critique, revise, and exchange intermediate conclusions. Finally, the strongest overall performance emerged in the multimodal dataset, while unimodal results remained competitive, suggesting a natural architecture in which modality-specific TAP-GPT agents communicate to simulate collaborative clinical decision-making. Together, these properties support framing TAP-GPT not merely as a predictor, but as a reasoning module suited for multimodal, communicative diagnostic systems. By operating on shared evidence rather than only probability scores, such a system would more closely resemble multidisciplinary clinical decision-making in the real world and provide a practical foundation for collaborative, multimodal LLM-based diagnostic support.

\section{Conclusion}
\label{sec:Conclusion}

In this work, we introduced TAP-GPT, a tabular large language model framework for few-shot Alzheimer's disease prediction on multimodal biomarker and imaging-derived data, and demonstrated that tabular-specialized LLMs can effectively reason over structured clinical data without serialization. Across multimodal and unimodal datasets, TAP-GPT consistently improved upon its backbone models, outperformed classical machine learning baselines, and remained competitive with dedicated tabular foundation models in low-shot settings, while additionally providing structured, clinically meaningful reasoning. We showed that TAP-GPT maintained robustness under missingness, generated structured and interpretable multimodal reasoning, and showed stability under self-reflection. TAP-GPT highlights a new direction for integrating multimodal biomarkers through structured LLM reasoning, laying the groundwork for future multi-agent diagnostic systems in which modality-specific expert models collaborate to produce more reliable and transparent clinical decisions.


\section*{Acknowledgments}

This work was supported in part by the NIH grants U01 AG066833, P30 AG073105, U01 AG068057, U19 AG074879, U01 AG088658, R01 LM013463 and R01 AG071470; and the NSF grant SCH 2500343. 


%


Data collection and sharing for this project were funded by the Alzheimer's Disease Neuroimaging Initiative (ADNI) (National Institutes of Health Grant U01 AG024904) and DOD ADNI (Department of Defense award number W81XWH-12-2-0012). ADNI is funded by the National Institute on Aging, the National Institute of Biomedical Imaging and Bioengineering, and through generous contributions from the following: AbbVie, Alzheimer's Association; Alzheimer's Drug Discovery Foundation; Araclon Biotech; BioClinica, Inc.; Biogen; Bristol-Myers Squibb Company; CereSpir, Inc.; Cogstate; Eisai Inc.; Elan Pharmaceuticals, Inc.; Eli Lilly and Company; EuroImmun; F. Hoffmann-La Roche Ltd and its affiliated company Genentech, Inc.; Fujirebio; GE Healthcare; IXICO Ltd.; Janssen Alzheimer Immunotherapy Research \& Development, LLC.; Johnson \& Johnson Pharmaceutical Research \& Development LLC.; Lumosity; Lundbeck; Merck \& Co., Inc.; Meso Scale Diagnostics, LLC.; NeuroRx Research; Neurotrack Technologies; Novartis Pharmaceuticals Corporation; Pfizer Inc.; Piramal Imaging; Servier; Takeda Pharmaceutical Company; and Transition Therapeutics. The Canadian Institutes of Health Research is providing funds to support ADNI clinical sites in Canada. Private sector contributions are facilitated by the Foundation for the National Institutes of Health (www.fnih.org). The grantee organization is the Northern California Institute for Research and Education, and the study is coordinated by the Alzheimer's Therapeutic Research Institute at the University of Southern California. ADNI data are disseminated by the Laboratory for Neuro Imaging at the University of Southern California. 



\section*{References}

\bibliographystyle{IEEEtran}
\bibliography{main}

\end{document}